\documentclass[review]{elsarticle}

\journal{Journal of \LaTeX\ Templates}
\usepackage[english]{babel}
\usepackage{float}
\usepackage{amsmath}
\begin{document}

\begin{frontmatter}

\title{Concept Learning through Deep Reinforcement Learning
with Memory-Augmented Neural Networks}

%\author{Jing Shi,Jiaming Xu\fnref{myfootnote},Yiqun Yao,Bo Xu}

\author[d1,d2,d3]{Jing Shi}

\author[d1,d2]{Jiaming Xu\corref{mycorrespondingauthor}}
\cortext[mycorrespondingauthor]{Corresponding author}
\ead{jiaming.xu@ia.ac.cn}

\author[d1,d2,d3]{Yiqun Yao}
\author[d1,d2,d3,d4]{Bo Xu}

\address[d1]{Institute of Automation, Chinese Academy of Sciences (CASIA), Beijing, China}
\address[d2]{Research Center for Brain-inspired Intelligence, CASIA}
\address[d3]{University of Chinese Academy of Sciences}
\address[d4]{Center for Excellence in Brain Science and Intelligence Technology, CAS, China}

\begin{abstract}
Deep neural networks have shown superior performance in many regimes to remember familiar patterns with large amounts of data. However, the standard supervised deep learning paradigm is still limited when facing the need to learn new concepts efficiently from scarce data.
In this paper, we present a memory-augmented neural network which is motivated by the process of human concept learning. The training procedure, imitating the concept formation course of human, learns how to distinguish samples from different classes and aggregate samples of the same kind. In order to better utilize the advantages originated from the human behavior, we propose a sequential process, during which the network should decide how to remember each sample at every step. In this sequential process, a stable and interactive memory serves as an important module. We validate our model in some typical one-shot learning tasks and also an exploratory outlier detection problem. In all the experiments, our model gets highly competitive to reach or outperform those strong baselines.
\end{abstract}


\end{frontmatter}

%\linenumbers

\section{Introduction}

The development of machine learning has brought great progress in the field of artificial intelligence. In particular, with extensive practice of neural network methods, many domains reach a new high, and, to our delight, some tasks have achieved the level of human beings. For example, neural machine translation (NMT) is so successful that for some language pairs it approaches the quality of human translators~\cite{1_wu2016google's}. Speech recognition technology has achieved a word error rate (WER) of only 5.9\%, which is similar to what human transcribers are able to achieve~\cite{2_xiong2016achieving}. In the field of computer vision, many works have been done to eclipse the abilities of people to classify objects defined in the ImageNet challenge~\cite{he2015delving}. However, despite those breakthroughs in the applications with deep neural networks, for the purpose of artificial general intelligence, it is still very far away. One setting that presents a persistent challenge is that of one-shot learning or small-sample learning.

Humans exhibit a strong ability to acquire and recognize new patterns. In particular, when presented with stimuli, people seem to be able to understand new concepts quickly and then recognize variations on these concepts in future percepts~\cite{lake2011one}. The process of learning new concepts by human is so efficient that we may utilize or imitate some mechanisms in it to expand the capacity of neural networks and fill the gaps with human intelligence.

Two approaches to learn concept in psychology are called concept formation and concept assimilation respectively. Concept formation means the method that key characteristics of things of the same kind can be discovered by learners from the various examples of similar things. Concept assimilation means the way that learner uses the existing concepts in the cognitive structure to understand the new concept. To put it in the context of machine learning, concept formation is akin to supervised learning, which recognizes a pattern after training of abundant data. Unfortunately, most successful neural networks are based on feeding of hundreds of examples and could only achieve good performance in the test dataset which is independent identically distributed with the training one. When new data is encountered, these models must inefficiently relearn their parameters to avoid the catastrophic performance~\cite{santoro2016one-shot}. Concept assimilation is similar to the scenario that the approach trained from the training dataset is able to learn new patterns quickly and even directly recognize them as new concepts. And in some ways, concept assimilation is also similar to meta-learning, learning to how to learn patterns.

In this paper, we present a novel neural network with memory module which is motivated by the process of human concept learning. The training procedure, imitating the concept formation of human, learns how to distinguish samples from different labels and to aggregate samples of the same kind. In test phase, the network will face new samples which have no overlap with the training dataset in both samples and even labels. In order to better utilize the advantages originated from the human behavior, we propose a sequential process for our network to decide how to remember the sample at every step. In this sequential process, a stable and interactive memory serves as an important module.	

We demonstrate that our model is capable of learning concepts from the frequent patterns and also mastering the method to represent the unknown ones which have never been seen. We evaluate our model on the well-known one-shot learning task Omniglot, which is the dataset with explicit one-shot learning evaluation~\cite{Kaiser2017Learning}. In this character-level visual concepts learning task, we get similar performance with the highly competitive works. Additionally, we evaluate our model on the speaker recognition task, which usually has a setup to test on the unknown speakers. Compared with those baseline methods, our model gets similar or better performance and shows gratifying adaptability. Finally, we explore the outlier detection ability of our model to test how to evaluate related works with a more human-like approach.

\section{Sequential Process to Form Concepts}\label{sec:seq}
Just like people study and learn as time and events go on, the whole process of our learning model is also based on a sequential process. At present, sequential process or sequence learning methods get much popularity in neural network based models~\cite{Graves2013Generating,Graves2013Speech,1_wu2016google's}. Take neural machine translation as an example: its architecture typically consists of two recurrent neural networks (RNNs), one to consume the input text sequence and one to generate translated output text~\cite{1_wu2016google's}. One of the benefits of sequential NMT lies in its ability to obtain history information when it makes prediction at every step. Compared with statistical machine translation, NMT sidesteps many brittle design choices such as word alignment, which usually produces a result by the original word alone but not in a semantic context.

Actually, for the task like one-shot learning, there have been some methods built on a setup of sequential process. In the paper from Santoro et al.~\cite{santoro2016one-shot} (hereinafter dubbed MANN), the authors present a form of episode for the model to learn the representation of each sample and also a mechanism to bind each sample to its class correctly. In one episode with length of $\emph{T}$, a dataset $D{\rm{ }} = {\rm{ }}\{ {d_t}\} _{t = 1}^T = \{ ({x_t},{\rm{ }}{y_t})\} _{t = 1}^T$ is given as a sequence. At each step $\tau$, the model, along with a memory module, gets the history subsequence $\{ ({x_t},{\rm{ }}{y_t})\} _{t = 1}^{\tau  - 1}$ and a single sample $x_\tau$  to make prediction about how to store it in the memory module.
Beyond that, Kaiser et al.~\cite{Kaiser2017Learning} present a long-life memory, and the whole learning process is based on this memory which stores representation of past samples and their labels.

In our work, we adopt a setup like the episode in MANN. This sequence-based setting, in our view, gets some advantages: (a) the history information could form various states, and these different states could make fully use of the training dataset through diverse permutations and combinations; (b) the sequential process is close to the way of human to form concepts and learn things, and this setting could better simulate this human-like behavior. In Section~\ref{sec:model}, the proposed setting would be further explained in detail.

\section{Related Work}

\subsection{One-shot Learning}
In recent years, research into one-shot or few-shot learning is fairly compelling and has received widespread attention by the machine learning community. Compared with the limited speed of development within the first decade, one-shot learning and some related works, such as meta-learning and transfer learning, have come into a new period.

The earliest and infrastructural work originates from the work by Li et al.~\cite{Fei2003A,feifei2006one-shot}. In those works, the authors developed a Bayesian framework to model the generation of data and to conduct one-shot image classification using the premise that previously learned classes can be leveraged to help forecast future ones when very few examples are available from a given class~\cite{koch2015siamese}. After a series of explorative works, Lake et al.~\cite{lake2011one} introduced a generative model of how characters are composed from strokes, where knowledge from previous characters helps to infer the latent strokes in novel characters. And they introduced the Omniglot dataset suitable for large scale concept learning from few examples, which has become a standard to test one-shot learning and related works.

Especially since 2015, several successful approaches for one-shot classification have been built and demonstrated good performance with specialized neural network architectures. Koch et al.~\cite{koch2015siamese} employed a Siamese network that was trained to rank similarity between inputs and predict whether two images belong to the same class. Vinyals et al.~\cite{vinyals2016matching} learned an embedding function and used cosine distance in an attention kernel to judge image similarity. Snell et al.~\cite{snell2017prototypical} employed a similar approach to Vinyals et al.~\cite{vinyals2016matching}, but used Euclidean distance with their embedding function~\cite{mishra2017meta}. Those works aim at building a more general function to get better representation of the dataset even when the data is from unknown distributions. After getting the representation, some metrics have been applied to calculate the similarities. Besides, as mentioned in the Section~\ref{sec:seq}, there has been several works~\cite{santoro2016one-shot,Kaiser2017Learning} using specific memory modules to save the class-dependent history information for the latter samples to decide which class is the most similar one.

Besides the common experiments and tests towards images, in our work, we make some attempts towards the speech field. In speech field, there have been some works~\cite{Fayek2017Evaluating} to test the dynamic classification problem toward kinds of tasks. Here we take the speaker identification problem as the point to test the learning of new concepts.

\subsection{Memory-Augmented Neural Networks}
In fact, neural networks based approaches were born with a very close relation with memory. For almost all classification tasks and advanced generation tasks, the networks are trained to form a set of fixed parameters to store the information of distribution in training dataset. That is to say, the parameters themselves are solid memories.

With the vigorous development of deep learning and neural networks, different memory-augmented models have been heavily studied. According to the different forms of existence, we conclude them into three types: built-in implicit memory, built-in explicit memory and long-life memory.

The built-in memory refers to the memory formed from current training samples (usually current mini-batch) while the long-life memory means a lasting memory module in the whole training process. The different kinds of recurrent neural networks are the typical built-in implicit memories. In recurrent networks, the state passed from one time step to the next can be interpreted as the network's memory representation of the current example. Moving away from this fixed-length vector representation of memory to a larger and more versatile form is at the core of these methods~\cite{Kaiser2017Learning}. Compared with the built-in implicit memory whose memory is the latent and inherent state of the network, built-in explicit memory gets a more definite structure that it usually declares a module of sequential vectors or a matrix as memory to store information. Content-based or position-based addressing algorithms are most common to load history information from these memory modules. Neural turing matching (NTM)~\cite{graves2014neural} and memory networks~\cite{weston2014memory} are representative models in this category. In NTM, Graves et al. augmented a recurrent neural network with a computing-inspired memory component that can be addressed via queries. Sukhbaatar et al.~\cite{sukhbaatar2015weakly} present a similar augmentation and show the importance of allowing multiple reads and writes to memory between inputs. These approaches excel at tasks where it is necessary to store large parts of a sequential input in a representation that can later be precisely queried. The end-to-end memory network~\cite{sukhbaatar2015end} is also of this kind. It presents a novel recurrent neural network architecture where the recurrence reads from a possibly large external memory multiple times before outputting a symbol. In addition, the MANN belongs to the built-in explicit memory, which varies according to different episode.

In the one-shot learning task, Kaiser et al.~\cite{Kaiser2017Learning} introduced a long-life memory module. This module enables one-shot learning in a variety of neural networks. As the network is trained, the memory increases and becomes more useful without the need to reset it during different training samples.

In addition, besides these two main tasks, there are some other works getting some relations with our proposed method. For example, Mirza and Lin ~\cite{Mirza2016Meta} proposed a meta-cognitive online sequential extreme learning machine for class imbalance and concept drift learning task. Moreover, Zhang el al.~\cite{Zhang2014Neural} proposed a novel iterative two-stage dual heuristic programming framework to solve the similar situation towards the discrete-time switched nonlinear system, which may provide some referential solutions for computation systems.

\section{Our Approach}\label{sec:model}
In our work, we use a sequence-based model to conduct the choices of different samples when facing different states. Our work is motivated by the process of human concept learning, so we design the model by the following intuitions:

\begin{itemize}
  \item Humans manage to remember objects from same class to form a definite and unified concept when the label is told.
  \item Which known concept to accommodate a new sample is mainly depending on the label of the new sample, even when the ¡®appearance¡¯ of it differs from the old ones completely.
  \item When a new sample is shown but without knowing its label, people prefer to cognize it with the experience of the known concepts. But it does not mean that just to bind the new sample to a former concept, and it may also sometimes form a totally new concept.
\end{itemize}
\begin{figure*}
  \centering
  \includegraphics[width=12cm]{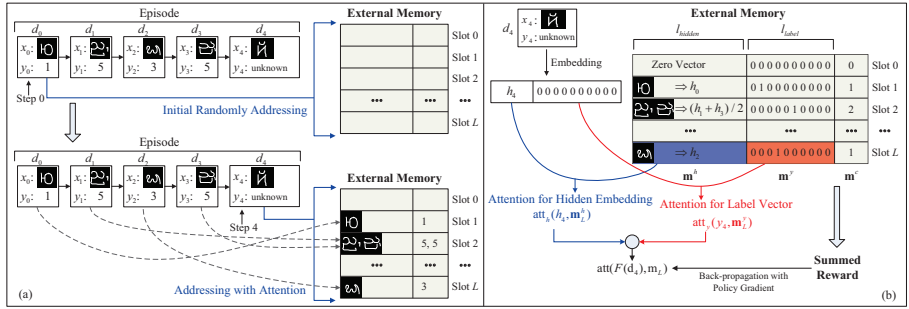}\\
  \caption{Illustration of one episode in our model. Part (a) shows two specific steps (first and last) during one episode, without knowing the label of the last sample. Part (b) describes the last step 4 in more detail with embedding function, the two-channel attention towards the \emph{L}-th slot and current state of the memory. The Summed Reward is calculated according to the state of memory after the action of final step. After getting the reward for this episode, policy gradient methods is used to train the network.} \label{fig:frame}
  %\vspace{-0.4cm}
\end{figure*}
\subsection{General Framework}\label{subsec:frame}
To put it simply, the whole process of our model seems like a neural clustering algorithm. The specialties lie in that the training process of our model is weakly supervised via some reward and punishment rules and the features of data are learned from the clustering process. As mentioned above, we adopt a setup like the episode in MANN~\cite{santoro2016one-shot}.  Consider an example in Figure~\ref{fig:frame}. The model within an episode is structured as
follows.
\begin{enumerate}
  \item There is a sequence of data samples represented by $D{\rm{ }} = {\rm{ }}\{ {d_t}\} _{t = 1}^T = \{ ({x_t},{\rm{ }}{y_t})\} _{t = 1}^T$, and $y_t$ is the class label for $x_t$. In the task of visual concept learning, $x_t$ could be the matrix of raw pixels of one image while $y_t$ could be a vector mapped from the label of $x_t$ with size of $l_{label}$. When the label of $x_t$ is unknown or not provided, we set $y_t$ to be a zero vector.
  \item A memory module $\bf{M}$ exists along with every step in this episode. Assume that the size of this memory is $\emph{L}$, which means $\bf{M}$ gets $\emph{L}$ slots to store the information during the episode. In one slot of $\bf{M}$, named $\bf{m}$, there are three elements: integrated hidden representations of samples ${{\bf{m}}^h}$, integrated labels vector ${{\bf{m}}^y}$, a counter  ${{\bf{m}}^c}$. The ${{\bf{m}}^h}$ is a vector of size $l_{hidden}$, which corresponds to the embedding size of sample $x$ in network while ${{\bf{m}}^y}$ is a vector of size $l_{label}$, which originates from the label $y$. And the counter ${{\bf{m}}^c}$ is an integer to record how many samples have been stored in this slot.
  \item The initial state of the memory $\bf{M}$ is empty. All the elements in this memory is set to be zero. And the first sample should enter the memory completely randomly through our algorithm.
  \item At step $\tau$, the network gets two inputs: memory $\bf{M}$ and current sample $({x_\tau },{\rm{ }}{y_\tau })$. Then, the network here first plays a role of embedding function $\emph{\textbf{F}}$ to embed the sample as follows:
      \begin{equation}\label{equ1}
      \emph{\textbf{F}}({d_\tau }) = \emph{\textbf{F}}({x_\tau },{\rm{ }}{y_\tau }) = Cont(\emph{\textbf{F}}({x_\tau }),{y_\tau }) = Cont({h_\tau },{y_\tau })
      \end{equation}
      In Eq. \ref{equ1}, in fact, the network only embeds the ${x_\tau}$  through its parameters and structure, so there exists $F({x_\tau }) = {h_\tau }$. In addition, $Cont$ means the operation to concatenate the hidden vector ${h_\tau}$ and the original label vector ${y_\tau}$. We unite the embedded sample $F({d_\tau })$ and the memory $\bf{M}$ to call them a state ${S_\tau}$. Facing this state, our model needs to take a policy $\pi ({S_\tau })$ to decide which slot of the memory should been chosen to store the hidden embedding $F({d_\tau })$ of this sample $({x_\tau },{\rm{ }}{y_\tau })$.
  \item Once the process is done, all the samples in this episode would be stored in the memory $\bf{M}$. And we could quantize the formation process through some standard. For instance, if all the samples clustered perfectly, which means each class in this episode has been merged into only one slot, then the reward should be very generous.
\end{enumerate}

\subsection{Memory Update}
Despite that kind of element-wised memory saving every single sample into one position, all the reused memories face a problem about how to update its content. Because there will be some samples to write into occupied positions.

Our work adopts a brief method to update the memory. As illustrated in Figure~\ref{fig:frame}(b), we only update one slot at every step. So the two crucial problems here are: (1) where to save the new sample, (2) how to write into the chosen slot.

For the former problem, we use a design to synthesize the similarity of hidden embeddings and label vectors. This method works as a two-channel attention network. Assume that ${{\bf{m}}_i}$  is the \emph{i}-th slot in memory. Then the policy $\pi ({S_\tau })$, also a probability function $\emph{\textbf{P}}$, decides which slot to store.

\begin{equation}\label{equ2}
\emph{\textbf{P}}(\emph{\textbf{F}}({d_\tau }),{{\rm{m}}_i}) = {\mathop{\rm Softmax}\nolimits} ({\mathop{\rm att}\nolimits}(\emph{\textbf{F}}({d_\tau }),{{\rm{m}}_i}))
\end{equation}

\begin{equation}\label{equ3}
%\begin{split}
{\mathop{\rm att}\nolimits} (\emph{\textbf{F}}({d_\tau }),{{\rm{m}}_i})) = {\mathop{\rm att}\nolimits} (Cont({h_\tau },{y_\tau }),{{\rm{m}}_i})\\
 = (1 - \lambda ) \cdot {{\mathop{\rm att}\nolimits} _h}({h_\tau },{\rm{m}}_i^h) + \lambda  \cdot {{\mathop{\rm att}\nolimits} _y}({y_\tau },{\rm{m}}_i^y)
%\end{split}
\end{equation}

\begin{equation}\label{equ4}
\lambda {\rm{ = }}{\mathop{\rm sgn}} (norm({y_\tau }))
\end{equation}

In the training phase, the model choose the slot as the obtained probability distribution while the test phase chooses the action with  largest probability.

In Eq. \ref{equ2} and Eq. \ref{equ3}, ${\mathop{\rm att( \cdot )}\nolimits}$  means the attention function to calculate the score between this sample and memory slot ${{\bf{m}}_i}$. And ${\mathop{\rm att_h( \cdot )}\nolimits}$  means the attention over the hidden embedding part while ${\mathop{\rm att_y( \cdot )}\nolimits}$  is the attention of the label vectors part. In Eq. \ref{equ3}, $\lambda$  is the switch to choose whether to attend via hidden embeddings or label vectors. In practice, we could meet some samples with or without knowing their labels. When a sample ${x_\tau}$ is provided with ${y_\tau}$, the network prefers to choose the slot whose label vector is closest to ${y_\tau}$. When the true label is not provided, we set ${y_\tau}$  to be zero vector, then the policy is executed totally based on the sample's content. This setup is quite natural because we human do assimilate a new sample with its homologous concept in memory directly. Assume that we have known some species of aves but excluding penguin, when a penguin is encountered and told as a kind of aves, we will expand this concept, although penguin is quite different with the former concept. We believe this ``supervised" process is crucial to form cognition and key features of one concept, especially when the concept is not so solid.

For the problem of how to save new sample, we update the chosen slot by taking average of all the history samples saved in it. And the counter of this slot will add one as a matter of course. This process could be concluded as follows.
\begin{equation}\label{equ5}
{\bf{m}}_i^y \leftarrow \frac{{{\bf{m}}_i^y \cdot {\bf{m}}_i^c + {y_\tau }}}{{{\bf{m}}_i^c + 1}},{\bf{m}}_i^h \leftarrow \frac{{{\bf{m}}_i^h \cdot {\bf{m}}_i^c + {h_\tau }}}{{{\bf{m}}_i^c + 1}}
\end{equation}
\begin{equation}\label{equ6}
{\bf{m}}_i^c \leftarrow {\bf{m}}_i^c + 1
\end{equation}

\subsection{Training Methods}
The sequence-based setup in our work provides a superexcellent scenario to use deep reinforcement learning (DRL) method. From above description, we could easily collect key elements to use the DRL algorithm.

In a DRL task, the network (as agent) would take action to change the state of environment and get corresponding reward. The action is based on policy $\pi ({S_\tau })$, which usually choose action according to the probability distribution in training and exploration phase and choose the action with maximum probability in test phase.

As mentioned in Section~\ref{subsec:frame}, in our work, the state could be set by combining $\emph{\textbf{F}}({d_\tau )}$ and memory $\bf{M}$. The policy $\pi ({S_\tau })$ is a two-channel attention model. After taking action to choose which slot to save at one step, the memory will be changed and the next sample will be loaded to form a new state. The reward is set on purpose to encourage to form a ``perfect" cluster, which means all the samples belonging to the same class will be aggregated into one slot without any error, and different classes get their own slots.

In our model, we use the REINFORCE algorithm~\cite{williams1992simple}, a method from policy gradient in reinforcement learning, to learn a policy by optimizing the long-term reward from ongoing sequential samples in an episode. The parameters gain gradient as follows.

\begin{equation}\label{equ7}
\begin{aligned}
{\nabla _\theta }J({\pi _\theta }){\rm{ = }}\int {_S{P^\pi }(a|s)} \int_A {{\nabla _\theta }}{\pi _\theta }(a|s){R^\pi }(s,a)dads \\=
{E_{s \sim {P^\pi },a \sim {\pi _\theta }}}[{\nabla _\theta }\log {\pi _\theta }(a\left| s \right.){R^\pi }(s,a)]
\end{aligned}
\end{equation}

Where $\emph{a}$ is the action to take (which slot to save), $R(s,a)$ is the reward when taking action $\emph{a}$ in state $\emph{s}$. $\emph{S}$ is the state space while $\emph{A}$ is the action space.\footnote{We refer readers to Williams~\cite{williams1992simple}  and Peters~\cite{peters2010policy}  for more details.}

To summarize, in one episode, the flow of the framework could be concluded as follows:
\begin{enumerate}
  \item \textbf{Start stage.} The first sample, faced with an empty memory, will be put into one slot randomly according to the equal probability over all the slots calculated from the attention function.
  \item \textbf{Middle stage.} In the following middle stage, the samples will be given iteratively. At every step, the attention function will output an probability distribution over all the slots to serve as the score for this current sample to choose. In training phase, the model takes the action according to the probability distribution. In test phase, the action with maximum probability will be chosen. After taking the action, an immediate reward at this step will be given by some rules based on the label of this sample and the states of the memory.
  \item \textbf{Final stage.} For the last sample in this episode, after the action of it, the final reward will be given according to the whole states of the memory. In the training phase, the final sample may face a chaotic memory while in test phase, owing to the known information about all the samples from start stage and middle stage, it usually faces a perfect cluster. More details about the process in test phase within one episode will be shown in following section about experiment.
\end{enumerate}
	
The setting of rewards is a bit tricky. At the very start, if one episode forms perfect clusters, we only set the final step of an episode to gain a large reward while the former steps are all set to 0. But we find this obvious preference get many difficulties in sampling the right path, especially when the length of the episode is long\footnote{Actually, the length larger than 5 is quite difficult to find. Because the number of candidate paths will increase exponentially.}. After kinds of attempts, we take a loose condition to initialize our model. Under this condition, we use the episodes with fewer kinds of labels (e.g. 2) and shorter length (e.g. 3) to pre-train the network so as to relieve the difficulty in finding the right path directly with longer episode. In the middle of one path, we only set punishment that, at every step, when the action puts the sample into an unused slot, we set the reward of this step to -1 to encourage the aggregation of the samples. When the action puts the sample into an used slot which has sample from different class with it, the reward will be -3. By contrast, at the final step, if the samples in this episode clusters perfectly, we set the reward to be a large positive score (in our experiments, we set it to be 100) to promote this selection of the path. Finally, We add up all the rewards from the beginning to the end of one episode without discount to serve as the scores for this episode with the chosen path.

The selection of rewards are based on some empirical observation. To be specific, the large score with the perfect result is not so steadfast. In the experiments, we find that the reward of 20 also works well in training with the short length (under 5). However, the large reward promotes the speed of converge. With the increase of the episode length, this phenomenon becomes more significant. Moreover, the negative reward in the middle stage is not necessary, but it will prevent the samples to aggregate together. At the beginning of the training process, this setting could avoid the frequent occupation of empty slots, which accelerates the training to a certain extent.

\section{Experiments}

In this section we describe the results of many experiments, comparing our model against many strong baselines.  Following the original intention from human concepts learning process, the experiments we set are like a way to test our model about the performance towards concepts learning efficiency, especially the new concepts. Based on this idea, the first two of our experiments revolve around the same basic task: a $N$-way $k$-shot learning task. Each method is provided with a set of $k$ labelled examples from each of $N$ classes that have not previously been trained upon. The task is then to classify a disjoint batch of unlabelled examples into one of these $N$ classes. Thus random performance on this task stands at 1/$N$. We compared a number of alternative models, as baselines, to our work.  Besides the popular and standard few-shot learning tasks,
%we also design a task to explore the reuse of our memory with some concepts stored in it.}.
 we also go further to analyze the zero-shot task with our model, attempting to explore the gap with real human intelligence.

We ran one-shot experiments on two data sets: Omniglot image classification set and Fisher speech dataset to learn the concepts about different speaker identities. It is worth to mention that the popular $N$-way $k$-shot learning tasks actually is  a nearest-neighbor classification method to bind the unlabelled new sample to its nearest single one. To be more specific, in the test phase, the $N$-way $k$-shot learning tasks is organised as follows. First, there are some new samples from $N$ unseen classes, and each class gets $k$ samples. Second, give this batch of data as pre-given samples, with their labels known, at a random order to one model to process them. And then, given a totally new sample, without the information of its label, from one of the $N$ unseen classes, the model should output the most likely class of this new sample by taking the label of the most similar sample as its. Repeate this process for thousand of times (with different batch of data) to calculate the final $N$-way $k$-shot accuracy. This method is adopted widely in many works such as~\cite{vinyals2016matching},~\cite{koch2015siamese}, ~\cite{lake2011one} and ~\cite{snell2017prototypical}. Using this setting, the $N$-way $k$-shot learning accuracy is depend entirely on the embedding of the single samples.

Different from the nearest single sample, memory-augmented methods may involve fusion of various samples, the classification basis is not just comparing similarities between single samples but clusters. In our work, during the test phase, the pre-given samples will be stored into the memory one-by-one. Owing to the known information about their labels, this storage process is quite confident to form perfect clusters. And the final sample with no information about its class will be classified into the most similar slot and gets its predicted label. Compared with those methods to judge according to single samples, this setting may bring more difficulties but also better scalabilities.

\subsection{Basic Attention Methods}
The aim of our model is to build an unified model with adaptability and flexibility. The flexibility lies in that different tasks could equip different networks as their embedding functions with experience and prior knowledge. Beyond that, the attention algorithm reflects the uniformity.

In all our model, attention function $att$ is used to calculate the similarity between two vectors. For the hidden embedding attention function $att_h$, we did a lot of attempts. From parametric approaches, we tested some typical attention functions which mainly used in natural language related neural networks, such as $h{\rm{W}}m_i^h$, ${{\rm{W}}_{\rm{1}}}h{\rm{ + }}{{\rm{W}}_{\rm{2}}}m_i^h$ and ${{\rm{W}}_3}\tanh {\rm{(}}{{\rm{W}}_{\rm{1}}}h{\rm{ + }}{{\rm{W}}_{\rm{2}}}m_i^h)$. From non-parametric methods, we tested dot production, cosine similarity and Euclidean distance. Unfortunately, in our experiments, we find the parametric approaches are very hard to train through RL frameworks. And we find Euclidean distance is the best in the non-parametric methods. So we finally choose Euclidean distance as our $att_h$  function in all our tasks. We think these difficulties in parametric methods may due to the specificity about the training process in one-shot learning task. To be specific, there gets no any fixed training dataset but massive episodes sampled from the random permutation of different known samples. The network with parametric attention methods may be led by the gradients to focus on the last attention parameters rather than the basic feature extraction part. And unfortunately, the massive randomly sampled episodes aggravate the difficulties in training the feature extraction network. In contrast, non-parametric methods gets better generalization by concentrating on the basic networks to extract some common hidden embedding from those images.

For the label vectors attention function $att_y$, we want to build an algorithm to identify whether two label vectors are same or not. In many memory-augmented methods, this judgment is crucial, but most of them are rule-based and not differentiable. In our work, we used a RNN to measure consistency between two label vectors. To be specific, we used one standard Gated Recurrent Unit (GRU)  layer and a fully connected layer as our attention function, which gets the difference value between two label vectors as input and outputs a scalar. We chose the RNN architecture because it could also be used with an extended input vector, which means the trained model could also adapt to another task even the new task get much more classes of samples (so label vectors may be longer).

We tested this idea through a simple experiment. First, we set the number of training labels to 10 and each of them is transformed to an one-hot vector with size of 10. Then we generated sequences with those vectors and trained the RNN based attention function $att_y$ through our framework till convergence. Finally, we tested the 5-way 1-shot accuracy of new label vectors. The new labels got a total number of 30,000 and we transformed them in to vectors with size of 15 based on their binary formats. The 100\% accuracy of this experiment proved that our design is efficient. And more important, it guarantee the almost complete accuracy when the label of the samples is given.

\subsection{Character-level Visual Concepts Learning}\label{exp1}
Omniglot consists of 1,623 characters from 50 different alphabets. Each of these was hand drawn by 20 different people. The large number of classes (characters) with relatively few data per class (20) makes this an ideal data set for testing small-scale one-shot classification~\cite{vinyals2016matching}.  Following ~\cite{vinyals2016matching} and ~\cite{Kaiser2017Learning}, we resized all the images to 28$\times$28 and we augmented the data set with random rotations by multiples of 90 degrees and used 1,200 characters for training, and the remaining character classes for evaluation. By our experimental comparison with the original data set, the rotation operation prevents the network from overfitting remarkably. That the totally same network and setting result in an obvious one-shot accuracy gap, from less that 75\% to almost one hundred percent.

We used a simple yet powerful CNN as the embedding function in this task - consisting of a stack of two modules. The first one is two 3$\times$3 convolution layers with 128 filters, a batch normalization~\cite{ioffe2015batch}, a ReLU non-linearity~\cite{glorot2011deep} and 2$\times$2 max-pooling. The second module is much like the first one, except that the number of filters is 256. Then, two fully connected layers with dimension 300, a ReLU activation function and an intercalary batchnorm layer are used to form the final embedding. The whole CNN network serves as the embedding function $\emph{\textbf{F}}$  in this task. The output of the last embedding layer is used as stimulation to our memory module and the label of the nearest slot returned by the memory is used as the final network prediction.
\begin{table}[htb] %
\caption{Few-shot accuracies (\%) in setup of episode. The results of baseline models are from~\cite{santoro2016one-shot}.}\label{tb1}
\begin{center}
\begin{tabular}{ccccccc}\hline
\textbf{Model}& ~1-shot~  & ~2-shot~ & ~3-shot~ & ~4-shot~\\\hline
Human &57.3 &70.1 &71.8 &81.4\\
FeedForward &19.6 &21.1 &19.9 &22.8 \\
LSTM &49.5 &55.3 &61.0 &63.6 \\
MANN &82.8 &91.0 &92.6 &94.9 \\\hline
Our Model &\textbf{96.6} &\textbf{98.6} &\textbf{99.3} &\textbf{99.6} \\\hline
\end{tabular}
\end{center}
%\vspace{-0.6cm}
\end{table}
Due to the similar episode-based sequence structure, we first compare our work to the MANN with totally the same setup and evaluation method. After training from episodes whose length is no more than 15, we test our model on episodes with length of 50. In each episode, 5 unique classes are randomly chosen from dataset. At every step of the episode, our model is used to output the prediction label, and this results from thousands of episodes are collected to form the evaluation result. Results comparing the MANN and baselines to our model are shown in Table~\ref{tb1}.

For all the few-shot accuracy, our model outperforms the baselines. We think the advantages may originate from several aspects. For example, the CNN-based model used in our work may better extract the features from the images rather than the LSTM used in MANN. In addition, the clusters formed in the memory slots are quite stable, which may avoid the influence from some special samples. Beyond these differences, both our model and MANN manage to learn not only the embedding function of samples but also the operation to bind the new sample with the memory component. And this operation is achieved by Kaiser et al.~\cite{Kaiser2017Learning} through man-made rules.

\begin{table*}[htb] %
\caption{5-way and 20-way, 1-shot and 5-shot classification accuracies on Omniglot, with 95\% confidence intervals where reported. For each task, the best-performing method is highlighted.}\label{tb2}
\begin{center}
\setlength{\tabcolsep}{0.6mm}{
\begin{tabular}{|c||c|c|c|c|}\hline
%\textbf{Model}& ~5-way 1-shot (\%)~  & ~5-way 5-shot (\%)~ & ~20-way 1-shot (\%)~ & ~20-way 5-shot (\%)~\\\hline
\textbf{Model}& ~5-way~  & ~5-way~ & ~20-way ~ & ~20-way ~\\
& ~1-shot (\%)~  & ~5-shot (\%)~ & ~1-shot (\%)~ & ~ 5-shot (\%)~\\\hline
Koch et al.~\cite{koch2015siamese} &97.3 &98.4 &88.2 &97.0 \\
Vinyals et al.~\cite{vinyals2016matching} &98.1 &98.9 &93.8 &98.5 \\
Finn et al.~\cite{finn2017model} &98.7$\pm$0.4 &\textbf{99.9$\pm$0.3} &95.8$\pm$0.3 &98.9$\pm$0.2 \\
Snell et al.~\cite{snell2017prototypical} &97.4 &99.3 &96.0 &98.9 \\
Munkhdalai and Yu~\cite{munkhdalai2017meta} &98.9 &- &97.0 &- \\
Kaiser et al.~\cite{Kaiser2017Learning} &98.4 &99.6 &95.0 &98.6 \\
Mishra et al.~\cite{nikhil2017meta}  &98.96$\pm$0.20 &99.75$\pm$0.11 &\textbf{97.64$\pm$0.30} &99.36$\pm$0.18 \\\hline
Our Model &\textbf{99.09$\pm$0.38} &99.65$\pm$0.22 &97.05$\pm$0.15 &\textbf{99.37$\pm$0.10} \\\hline
\end{tabular}}
\end{center}
%\vspace{-0.6cm}
\end{table*}
Furthermore, we compare our model with many strong approaches in more popular $N$-way $k$-shot tasks. Compared with the setup in MANN, this task focus more on building an embedding method with better generalization. To concentrate on the embedding methods only, we set all the training samples with zero label vectors except the first sample of one specific class (named seed samples) during each episode. This set forces samples to be clustered only through the similarities between their content but not the labels. Results comparing the baselines to our model on Omniglot $N$-way $k$-shot tasks are shown in Table~\ref{tb2}.

For both 1-shot and 5-shot, 5-way and 20-way, our model gets comparable performance with the highly competitive methods. This result proves that different classes in test data set are well-distributed in the hidden space through our network. Samples from same class tend to gather and different class tend to be far away from each other.
%However, 5-shot results are relatively not so good as the 1-shots¡¯. This may be originates from that our prediction is made according to cluster rather than single nearest sample. Some widely distributed class may have some

\subsection{Concept Learning of Speaker Recognition}

From the beginning of our work, we plan to build an unified framework to strengthen ability to learn kinds of different concepts. Thus, besides the widely adopted Omniglot visual dataset, we also did some experiments towards speaker recognition task.

The Fisher database was used in our experiments. The training and evaluation datasets are presented as follows.
\begin{itemize}
  \item \textbf{Training set}: It consists of 2,500 male and 2,500 female speakers, with 95,167 utterances randomly selected from the Fisher database, and each speaker has about 25 seconds speech segments on average. After extracting 40-dimension F-bank features from this data, we split each segments to samples with 40 frames (0.4 second). That means all the samples get a size of 40$\times$40, all the 5,000 speakers get 3,306,995 samples in total. Finally, these features are used for training the widely used i-vector  system as baseline, and another Softmax classification network is also used to compare our model. The softmax model has an identical CNN embedding structure with our model and an additional softmax layer to class the training data.
  \item \textbf{Evaluation set}: It consists of 350 male and 350 female randomly selected speakers. There is no overlap between the speakers of the training set and evaluation set. But all the operations are identical to training set.
\end{itemize}

In detail, we use the same CNN-based network to serve as the embedding function without changing any setup. Further, the attention network $att_y$ used over label vectors is directly transplanted from the trained RNN network in Omniglot experiment without any adjustment. The only difference is that the length of the label vectors increase to 15 from 10

After training this speaker network, we also use the $N$-way $k$-shot accuracy used in Omniglot to compare different methods. The result is shown in Table \ref{tb3}.
\begin{table}[htb] %
\caption{ 5-way 1-shot and 5-shot evaluation accuracies (\%) in Fisher dataset. The entire i-Vector system was trained using the MSR Identity Toolbox. We tested many non-parametric metrics to get the accuracies and report the best one with cosine similarity. }\label{tb3}
\begin{center}
\begin{tabular}{ccccccc}\hline
\textbf{Model}& ~5-way 1-shot~  & 5-way~5-shot~ \\\hline
i-Vector &23.1 &27.3 \\
Softmax Network &75.5 &77.6  \\\hline
Our Model &\textbf{83.0} &\textbf{88.4} \\\hline
\end{tabular}
\end{center}
\vspace{-0.6cm}
\end{table}

Compared with the most advanced results in speaker identification task with accuracy over 90 percent, our results seems not so strong. This gap gets kinds of reasons: (1) the number of enrollment ($k$ in $k$-shot) is very restricted in our model, because we want to evaluate the performance over very few new samples. (2) The duration of one sample to learn the embedding function is very short that we get only 0.4 second rather than 3$\sim$10 seconds in many other works~\cite{li2017deep,li2017deep1}. (3) The total time of a speaker is also very restricted. (4) The embedding network we used is not specially designed but the same structure with Omniglot task, which may be better suitable for images.

However, even with a relatively small scale, the trends are well reflected. The results of our model also show obvious superiority over baseline models. Compared with the Softmax network, though the embedding method is same, our model could use the relations between different samples to train an embedding function with better generalization ability. This proves once again that our framework gets the ability to learn a well embedding method for different tasks.

\subsection{Outlier Detection Test}

In Section~\ref{exp1}, we allocate label vectors of the seed samples to guide the model to focus on completing the $N$-way $k$-shot tasks. Although this trick is useful, it may avoid the important problem about outlier detection (in MANN, this problem is called zero-shot learning). Actually, in almost all the one-shot learning works, this problem has been sidestepped. However, this ability deserves to be discussed.

Humans are so intelligent that we could not only recognize newly acquired concept but also determine whether a sample is a totally new concept or not. Mapping this task to Omniglot dataset, the zero-shot learning means sometimes the model should judge a sample to belong to none of the provided classes. In our framework, zero-shot learning means storing a sample with unseen class successfully into an empty slot rather than an occupied one.

%\begin{figure}[htb]
%    \centering
%   \includegraphics[scale=0.5,bb=0 0 385 567]{222.png}
%   \caption{Some description about the picture}
%   \label{picture-label}
%   \end{figure}

In fact, the episode setup in our work gets a natural circumstance to conduct zero-shot classification. In each episode, there are certain samples whose label has never shown before. At these steps, a well-trained network should be able to put these samples into an empty slot in memory without knowing their label vectors. Actually, in our general framework, our model gets the aim to finish both zero-shot and few-shot tasks. But in practice, we find the few-shot abilities and zero-shot get some kinds of contradiction.
\begin{figure}[htb]
  \centering
  \includegraphics[width=8.5cm]{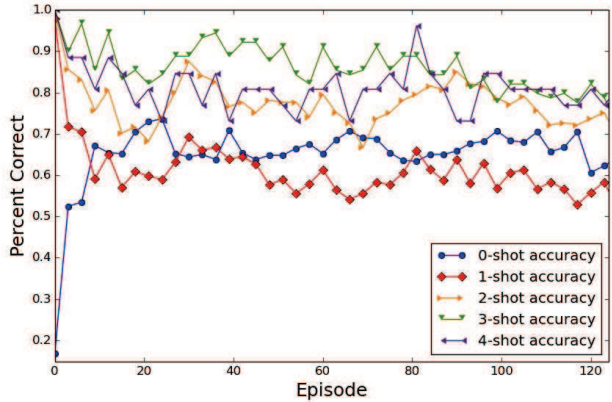}\\
  \caption{The curve to show the changing of accuracies when applying the well-trained few-shot network to finish zero-shot task. Horizontal axis means the number of episodes to update.}\label{fig:0shot}
  %\vspace{-0.6cm}
\end{figure}

As shown in Figure~\ref{fig:0shot}, we use the well-trained parameters in few-shot tasks to initialize the networks in zero-shot tasks. In zero-shot task, all the labels of samples are not provided, so the whole policy to choose the slot is only determined by the content of the samples. Just at the beginning, the zero-shot accuracy increases dramatically and the accuracies of few-shot learning decrease with similar trends. There are no major surprises in these results: to avoid choosing a wrong used slot, the network prefer to put most samples into an empty slot, this tendency results in an obvious improvement of zero-shot accuracy and deterioration in few-shot tasks. Finally, after competition between these two directions, the network reaches a trade-off.

\begin{table}[htb] %
\caption{5-way 0-shot, 1-shot accuracies (\%) and F1 score of them in Omniglot dataset. The few-shot of our model means the well-trained network from Section~\ref{exp1}. }\label{tb4}
\begin{center}
\begin{tabular}{cccc}\hline
\textbf{Model}& ~0-shot~  &1-shot~ & F-score \\\hline
MANN &36.4 &82.8 &50.6\\
Kaiser et al.~\cite{Kaiser2017Learning} &4.9 &98.4 &9.3 \\\hline
Our Model (few-shot) &16.8 &\textbf{99.1} &28.9\\
Our Model (zero-shot) &\textbf{68.4} &62.5 &\textbf{65.3}\\\hline
\end{tabular}
\end{center}
%\vspace{-0.75cm}
\end{table}
To compare different methods quantitatively, we calculated the F1 score from zero-shot and one-shot accuracy. F1 score is the harmonic average of the precision and recall used in statistical analysis, where an F1 score reaches its best value at 1 and worst at 0. Here we use this measure to test the performance for different models to handle both the 1-shot learning and outlier detection (0-shot learning).
%The one-shot accuracy shows a basic tendency of all the few-shots performance.
As shown in Table~\ref{tb4}, our model finds a better way to balance the two directions than other memory-augmented models.

\subsection{Error Analysis}
After getting the trained model, we could find that there are some typical unsuccessful situations deserved to be discussed.
%Here we will take the learning of visual concepts as example, which is more intuitive to observe.
In our experiments, compared with other works equally, the labels of pre-given samples before the final stage are all known. If we set the whole process in an episode decided by the content of the samples without knowing the label information, more problems could be observed. During one episode, there is unlikely to make wrong action in the beginning. Because every slot is empty and the network calculates the score over each slot and gets the exactly same probability for all the slots. After the storage of the initial sample, the following actions may bring some risks.
After the observation of the wrong actions, we find that the most frequent situation lies in that the appearance of a new class.
When a sample is the first from one specific class in this episode (but not the first one of the whole episode), its right action is to take one empty slot. However, with a certain probability, it may get a higher score over the existing samples stored in memory compared with the zero vectors in those empty slots, although the score is actually very small. If this wrong action happens, the following steps would be messed easily. Actually, this typical problem is easy to be fixed to set some hand-written rules, such as setting a threshold. However, for our framework, we construct the whole process with algorithm itself. From the viewpoint of human behavior, the whole brain learns new concept quickly and precisely without any higher guide. Different from some other works, we manage to weaken the influence of artificial work, although it may bring some errors.

\section{Discussion and Future Work}
In this paper we presented a new memory-augmented neural network architecture that can be used for learning simple concepts. There are some main contributions in our work. Firstly, at the modelling level, we combine the machine learning classification problems with the human concept learning process, inspired by which we set up a sequential process to direct the methods of how to learn embedding of samples with better generalization. Secondly, we use a memory module to simulate a diverse environment and we also provide a deep reinforcement learning method to successfully train our model. With this module, different classes of concepts get more liable to be away from each other and the samples of same kind tend to gather, which improves the storage efficiency. Further, we explore a new task to measure few-shot tasks among related works.

Although the performance achieves the level approximating the highly competitive methods, there still are some challenging problems. In this paper, each ``concept'' gets only a single form as image or voiceprint. In reality, most concepts are formed via different modes. We recognize a person through mixture of appearance, voice and even behavioral pattern. In future, we will try this multi-modal concepts formation task, hoping to contribute one step for further general intelligence.

%\section*{Acknowledgements}
%We thank the reviewers for their insightful comments and thank the editors for their conscientious working. This work was supported by the National Natural Science Foundation of China (61602479, 91720000), the Advance Research Program (6140452010101) and the Strategic Priority Research Program of the Chinese Academy of Sciences (XDBS01070000).

\section*{References}

\bibliography{icml2018ref_shin}

\end{document}